\documentclass[]{interact}

\usepackage{epstopdf}
\usepackage[caption=false]{subfig}

\usepackage{natbib}
\bibpunct[, ]{(}{)}{;}{a}{}{,}
\renewcommand\bibfont{\fontsize{10}{12}\selectfont}

\theoremstyle{plain}

\theoremstyle{definition}

\theoremstyle{remark}

\usepackage{url}
\usepackage{tabularx}
\usepackage{xspace}
\newcommand{\Xcal}{\mathcal{X}}
\newcommand{\XLabel}{\textsf{XLabel}\xspace}
\newcommand{\EBM}{\textsf{EBM}\xspace}
\newcommand{\ruleb}{\textsf{RuleBased}\xspace}
\newcommand{\XGB}{\textsf{XGB}\xspace}
\newcommand{\LGBM}{\textsf{LGBM}\xspace}
\newcommand{\SVM}{\textsf{SVM}\xspace}
\newcommand{\RF}{\textsf{RF}\xspace}

\begin{document}

\articletype{Research article}

\title{Developing A Visual-Interactive Interface for Electronic Health Record Labeling: An Explainable Machine Learning Approach}

\author{
\name{Donlapark Ponnoprat\textsuperscript{a}\thanks{CONTACT Donlapark Ponnoprat Email: donlapark.p@cmu.ac.th}, Parichart Pattarapanitchai\textsuperscript{a}, Phimphaka Taninpong\textsuperscript{a}, Suthep Suantai\textsuperscript{a}\thanks{CONTACT A.~N. Author. Email: latex.helpdesk@tandf.co.uk}, Natthanaphop Isaradech\textsuperscript{b} and Thiraphat Tanphiriyakun\textsuperscript{b}}
\affil{\textsuperscript{a}Data Science Research Center, Department of Statistics, Faculty of Science, Chiang Mai University, Chiang Mai 50200, Thailand; \textsuperscript{b}Sriphat Medical Center, Faculty of Medicine, Chiang Mai University, Chiang Mai 50200, Thailand}
}

\maketitle

\begin{abstract}
Labeling a large number of electronic health records is expensive and time consuming, and having a labeling assistant tool can significantly reduce medical experts' workload. Nevertheless, to gain the experts’ trust, the tool must be able to explain the reasons behind its outputs. Motivated by this, we introduce Explainable Labeling Assistant (XLabel) a new visual-interactive tool for data labeling. At a high level, \XLabel uses Explainable Boosting Machine (\EBM) to classify the labels of each data point and visualizes \emph{heatmaps} of \EBM's explanations. As a case study, we use \XLabel to help medical experts label electronic health records with four common non-communicable diseases (NCDs). Our experiments show that 1) \XLabel helps reduce the number of labeling actions, (2) \EBM as an explainable classifier is as accurate as other well-known machine learning models outperforms a rule-based model used by NCD experts, and 3) even when more than 40\% of the records were intentionally mislabeled, \EBM could recall the correct labels of more than $90\%$ of these records.
\end{abstract}


 \begin{figure}[!h]
\centering\includegraphics[width=\textwidth]{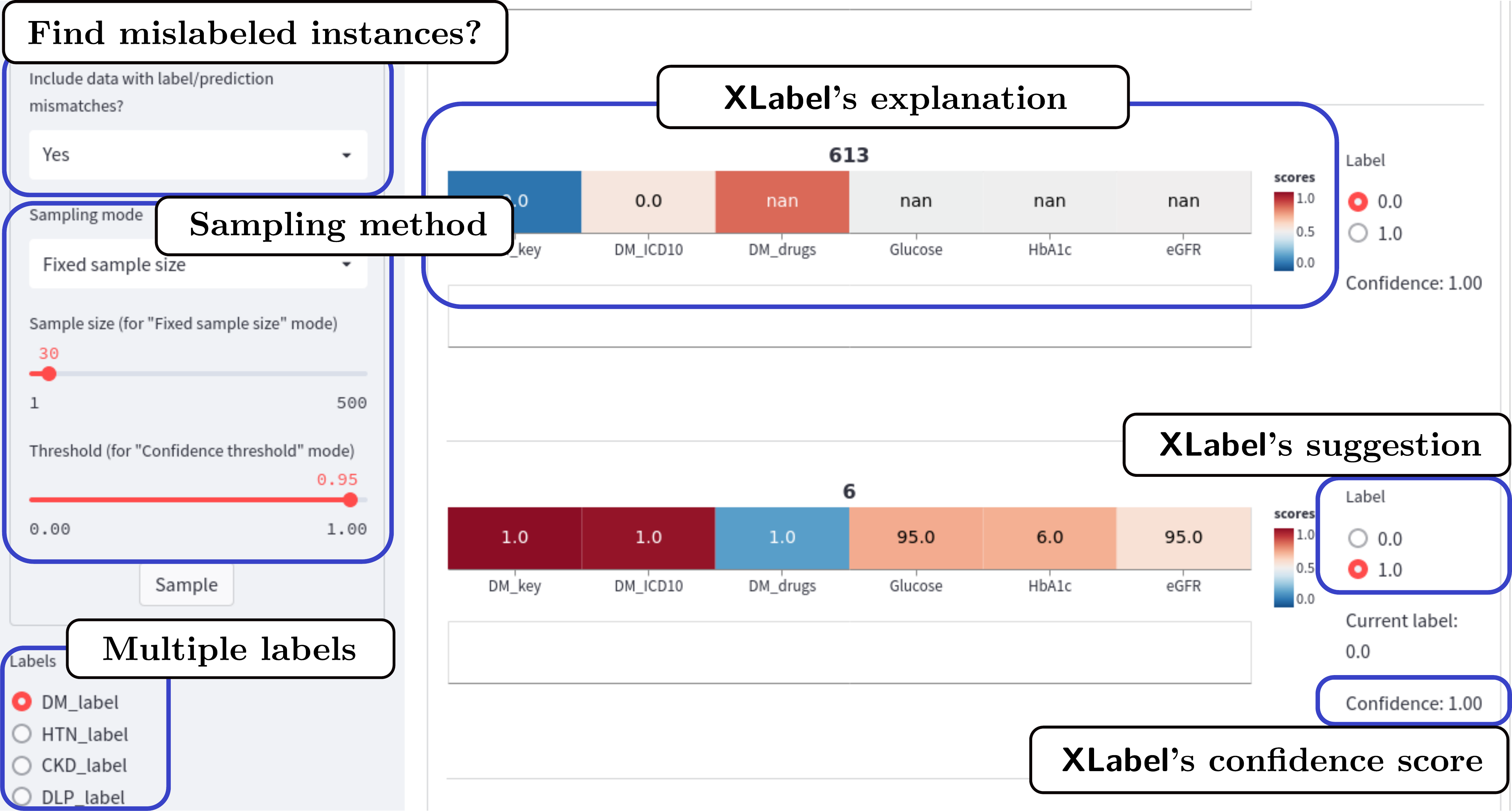}
\caption{The Explainable Labeling Assistant (\XLabel)~\label{fig:ss}}
\end{figure}   
\unskip

\section{Introduction}
In healthcare, there are many sources of every growing data, such as patients' records, electronic health records and laboratory results. If properly managed and analyzed, such data can provide useful information to patients, physicians and medical researchers, who then take advantage of the information to improve medical research and patient care. In spite of this, it will be difficult to conduct impactful research without a carefully labeled dataset. For example, an observational study of a particular disease will be difficult if each record in the dataset does not come with a clear disease label, which is common for records of follow-up visits. It then becomes a medical experts’ important task to label all electronic health records before releasing them to the public for future research use. 

Labeling a large amount of data can be expensive and time consuming. Thus, there has been increasing interest in using an assistant tool to speed up the labeling process. One approach to build such a tool is to treat the data labeling problem as a classification problem, where each input consists of each record’s features, and the output is the record’s label. Though such a problem can be tackled by machine learning (ML) models, many of these are black-box models, which cannot explain the internal processes that lead to their classification. Without explanations, the model has no grounds to convince the labeler that its classifications are the correct ones. 

\subsection*{Our Contributions} 

We present a novel visual-interactive tool called \XLabel, designed to enhance the data labeling process using an explainable machine learning (ML) approach. XLabel can:
\begin{enumerate}
    \item accurately predict the labels and provide a visual explanation of each input feature's influence towards the prediction,
    \item detect mislabeled data by comparing its predictions with the existing labels,
    \item in the case that a prediction is wrong, receive the user's correction and adjust the predictive model accordingly.
\end{enumerate}

\XLabel is thus an interactive human-machine system: its predictions and explanations reinforce the user’s labeling decisions, and new labels from the user allow \XLabel to improve the predictive model. In addition, \XLabel can also be used to detect mislabeled data by comparing its predictions with the existing labels.

As an application, we consider the task of labeling electronic health records of patients with potential non-communicable diseases (NCDs), which is one of the most concerning health issues worldwide. We will focus on four common NCDs: diabetes mellitus (DM), hypertension (HTN), chronic kidney disease (CKD), and dyslipidemia (DLP). We will design a user interface that allows the user to interact with the model’s predictions. In addition, we will perform experiments which demonstrate that 1) our explainable model is as accurate as black-box ML models, and 2) it can recall most of the correct labels even when a sizable portion of the data has been mislabeled. 

\subsection*{Related Work} 

There have been many studies that apply ML techniques for classification of various NCDs; for example, hypertension~\citep{Ambika2020a, Ambika2020b}, diabetes mellitus~\citep{Pei2019,islam2020b}, stroke~\citep{rosado2019,Rajora2021} and asthma~\citep{Finkelstein2017,mali2022}. Some of these works take the explainable ML approach. For example,~\cite{Rashed2021} use Shapley Additive Explanations (SHAP)~\citep{Lundberg2017} to interpret the decisions of various ML models for chronic kidney disease diagnosis. \cite{Shafi2022} use DeepSHAP~\citep{Chen2019} to explain ML models' classifications of Alzheimer's disease.~\cite{Davagdorj2021} use DeepSHAP to explain classifications of multiple NCDs.~\cite{cheng2020} use Partial Dependence Plot~\citep{Friedman2001}, SHAP, Anchors~\citep{Ribeiro2018} and Accumulated Local Effects~\citep{Daniel2020} to explain classifications of multiple NCDs. In contrast to these works, in which the ML models are trained on fully labeled datasets, our work is the first to employ an explainable ML model to assist with data labeling.

There has been a surge of interest in human-machine interactive labeling.~\cite{Nadj2020} have categorized interactive labeling systems into five design principles.~\cite{Viana2021} extend this work by also analyzing their user interfaces.~\cite{YAKIMOVICH2021100383} provide an extensive review over many automatic data annotation strategies, with different levels of human involvement. For specific methods,~\cite{Desmond2021} design a labeling assistant that uses a semi-supervised learning algorithm for label suggestions.~\cite{Ashktorab2021} design a labeling interface that presents the labeler with a batch consisting of nearest neighbors of a random example; these neighbors are likely to share a label. All in all, one must be careful when designing an interactive labeling system, as an experiment by~\cite{Bondi2022} show that human judgment is biased towards the model's classifications. To this end, our method introduces one way of reducing the bias---by providing the labeler with explanations of the classifications.

\section{Materials and Methods} 

\subsection{The Data Labeling Task} 

We start with a raw database of check-up records of NCD patients. Each record contains the following individual information:  
\begin{itemize}
    \item Personal features that are age, sex, height, and weight. 

    \item Laboratory results such as blood sugar level and blood pressures. 

    \item International Classification of Diseases (ICD-10) codes of diagnosed diseases. 

    \item Doctor’s notes. 

    \item A list of prescribed drugs. 
\end{itemize}

In the case of minor visits (e.g., to refill the prescription), the record might not have some laboratory results and ICD-10 codes. 

To make the database useful for future NCD analysis, we ask a medical expert to label each record with four NCD labels: diabetes mellitus (DM), hypertension (HTN), chronic kidney disease (CKD), and dyslipidemia (DLP). In other words, the database will have four additional columns, each of which contains the labels of each NCD. 

In this work, we introduce a new visual-interactive tool that helps with data labeling, which will be described in the next section. 

\subsection{Visual-Interactive Labeling} 

\begin{figure}[t]
\centering\includegraphics[width=10.5 cm]{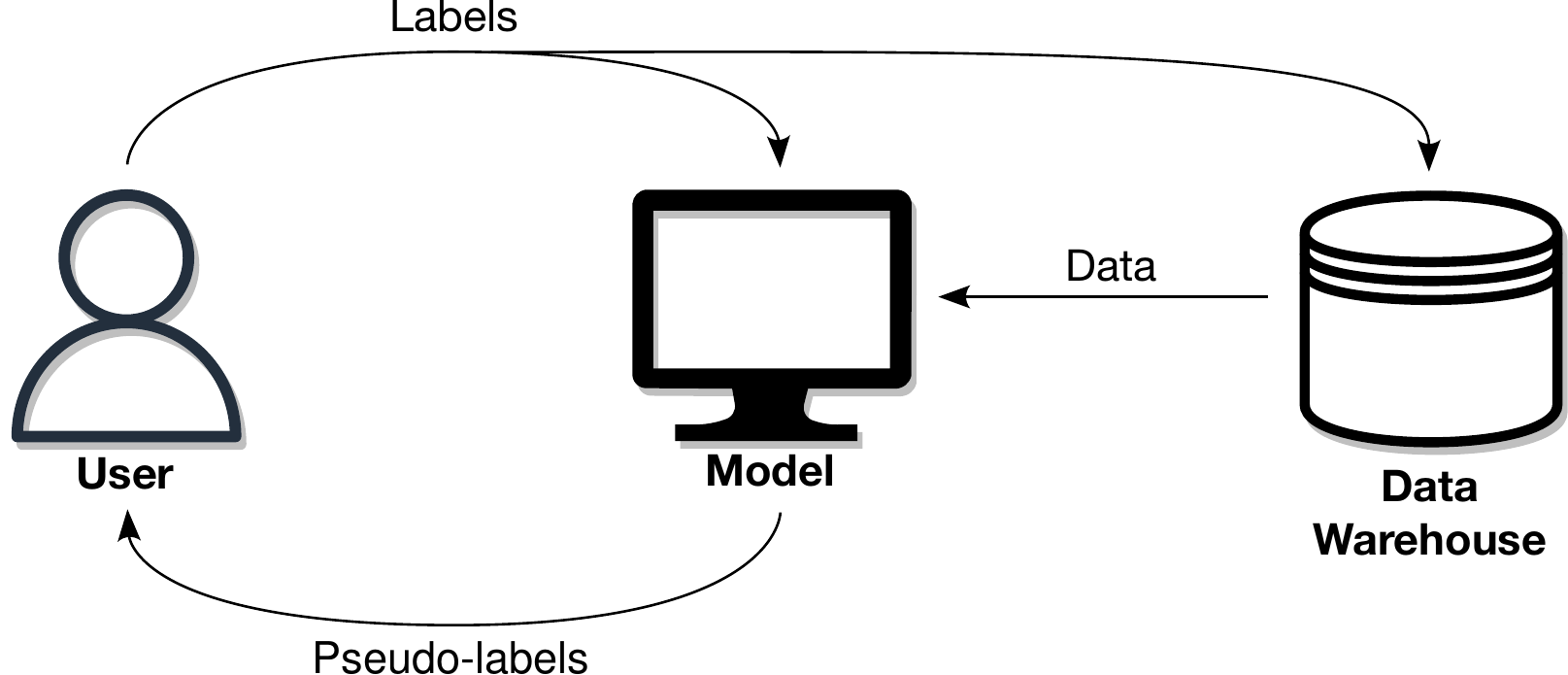}
\caption{A high-level picture of \XLabel. It sends pseudo-labels and their explanations to the user. The user then turns the pseudo-labels into true labels by keeping the correct pseudo-labels or flipping the wrong ones. The labels are then sent back to \XLabel and the data warehouse. \label{fig1}}
\end{figure}   
\unskip

Labeling massive medical data can be very time-consuming. To reduce medical experts’ workload, we design a visual-interactive tool called Explainable Labeling Assistant (\XLabel). The most important part of \XLabel is a classification model that takes a patient’s record as an input, and suggests a label $y\in \{0,1\}$ of that record to the user; here, $y$ is $1$ if the disease is present, and $0$ otherwise. 

To ensure that the model’s suggestions are trustworthy, we take an explainable approach; the model must be able to explain the reasons behind its suggestions. A high-level picture of the labeling process with \XLabel is as follows (also shown in Figure~\ref{fig1}): 
\begin{itemize}
    \item \XLabel reads the data of all unlabeled records, then creates a pseudo-label for each record. . 

    \item \XLabel shows a subset of records, their pseudo-labels, and their explanations to the user. 

    \item The user reads the explanations, then turns the pseudo-labels into true labels by keeping the correct pseudo-labels and flipping the wrong ones (i.e., from $0$ to $1$ or $1$ to $0$). 

    \item \XLabel accepts the labels from the user and retrains its classification model. Now the model can provide more accurate pseudo-labels to the next unlabeled sample. 
\end{itemize}
The user’s labeling workload will be vastly reduced if most of the pseudo-labels are already correct. Thus, in addition to being explainable, the classification model inside \XLabel must be accurate. Recently, there have been a series of works showing that, contrary to widespread belief that there is a trade-off between explainability and accuracy, it is possible for a ML model to be both explainable and accurate~\citep{Lipton2018,Rudin2019}. 

\subsection{Explainable Boosting Machine (\EBM)} 

The classification model that we use in \XLabel is Explainable Boosting Machine (\EBM)~\citep{Lou2013,nori2019}, an explainable version of gradient boosting machine~\citep{Friedman2001, Friedman2002}, which is known for its classification performance. 

Let $x=(x_1,\ldots,x_n)$ be a patient’s record with true label $y\in \{0,1\}$. The \EBM is an additive model, that is, its classification on $x$ is given by  
\begin{equation}\label{eq:ebm}
    f(x) = \beta_0 + \sum_i f_i(x_i) + \sum_{i\not= j} f_{ij}(x_i,x_j),
\end{equation}
where $\beta_0$ is the intercept, and each $f_i$ is a sum of regression trees, that is, 
\begin{align*}\label{eq:sum}
    f_i(x_i) &= \sum_{k} f_{ik}(x_i)\\
    f_{ij}(x_i,x_j) &= \sum_{k} f_{ijk}(x_i,x_j).
\end{align*}
Here, $f_{ik}$ and $f_{ijk}$ are regression trees for all $i$, $j$ and $k$. The model then outputs the class conditional probability through the logistic function: 
\begin{equation*}
    p_x = \Pr(y=1 \mid x) = \frac{1}{1+e^{-f(x)}}.
\end{equation*}
The classified label is then $\hat y = 1$ if $\Pr(y=1 \mid x)\geq 0.5$ and $\hat y = 0$ if $\Pr(y=1 \mid x)< 0.5$. 

\subsection{\XLabel’s Explanations} 

The fact that \EBM is an additive model allows \XLabel to measure the contribution from each input feature towards the classification. More precisely, from~\eqref{eq:ebm}, we can treat $f_i(x_i)$ as the contribution from $x_i$ (we shall ignore the interactive terms $f_{ij}(x_i,x_j)$ as they are used to model the residual~\citep{Lou2013}). In particular, $f(x_i)>0$ implies that $x_i$ contributes to a positive label, while $f(x_i)<0$ implies that $x_i$ contributes to a negative label.

To visualize these feature contributions, we chose the \emph{heatmap}, as its compact representation allows the user to scroll through the records very quickly. In each heatmap, a rectangle is drawn for each feature, and the color is determined by its contribution.

However, \EBM's feature contributions $f_{i}(x_{i})$ cannot be visualized as a color right away, as the value can be an arbitrarily large positive or negative number. So we propose to scale it to a range of $(0,1)$ using the logistic function: 
\begin{equation*}
    \operatorname{HEAT}(x_i) = \frac{1}{1+e^{-f_i(x_i)}}.
\end{equation*}
The rectangle is then colored red if $\operatorname{HEAT}(x_i)$ is close to $1$ and blue if it is close to $0$. This heatmap allows the user to quickly notice the features that contribute the most to the label, and then promptly decide to keep or flip the label. 

In addition to the heatmap, \XLabel displays the doctor’s notes and highlights keywords that are associated with the labels. The keywords are provided by the NCD experts. 

\subsection{Sampling Method}\label{sec:criteria}

Now, our goal is to make \EBM as accurate as possible with only a few labeled records sent from the user to \XLabel. To accomplish this, \XLabel sends the ``least confident'' records to the user. After the user submits the true labels, \EBM then learns from these labels and becomes more confident in classifying similar records. 

To compute the \EBM’s confidence score of a record $x$, i.e., how confident it is in its classification $\hat y$ of $x$, we use the \emph{misclassification} rate: 
\begin{equation}\label{eq:score}
    C_x = \min\{p_x, 1-p_x\}.
\end{equation}
Note that $C_x \in [0.5,1]$, $C_x= 1$ when \EBM is most confident in $x$’s classification (i.e., $p_x = 1$ or $p_x = 0$) and $C_x = 0.5$ when it is least confident (i.e., $p_x = 0.5$).

At the beginning, \XLabel lets the user choose between two sampling methods: 
\begin{itemize}
    \item \textbf{Confidence threshold}: \XLabel will select all records whose confidence scores are less than a threshold specified by the user. 

    \item $n$\textbf{-least confident}: \XLabel will select $n$ records with the smallest confidence scores. Here, the sample size $n$ is specified by the user. 
\end{itemize}
Starting from the class conditional probability $p_{x}$ of all records $x$, \XLabel computes the confidence scores according to~\eqref{eq:score} and samples a subset of records according to the chosen sampling method. 

\subsection{Correcting Mislabeled Data} 

Sometimes the expert might mislabel the data due to missing an important keyword or fatigue. For example, the expert might miss the ``DM'' tag (which indicates that the record has diabetes mellitus) in the clinical note and label the record as ``non-DM''. 

\XLabel can also be used to detect mislabeled records. To illustrate this, let us denote the whole dataset by 
$\Xcal=\Xcal_L\cup \Xcal_U$, where $\Xcal_L$ is the set of labeled records and $\Xcal_U$ is the set of unlabeled records. Suppose that the \EBM has been trained on $\Xcal_L$, which contains sufficiently many correctly labeled records. Instead of asking \EBM to classify only on $\Xcal_U$, \XLabel can ask \EBM to classify the whole dataset $\mathcal{X}$. Sometimes, there is a record whose \EBM’s classification is different from the current label, indicating that the record might be mislabeled; \XLabel will show such records (together with the sampled records from 
$\Xcal_U$ as described above) and ask the user to confirm or change the label.

\subsection{\XLabel's User Interface} 

We have implemented \XLabel in Streamlit (\url{https://streamlit.io}), which is an open-source application framework in Python. Inside \XLabel, we employ the InterpretML’s implementation of \EBM~\citep{nori2019}. A screenshot of the interface is shown in Figure~\ref{fig:ss}. The application is designed to support a wide range of tabular datasets, including those with multiple labels, each of which can have multiple classes. It also supports datasets with missing values. 

After the user uploads a file of unlabeled records, they will be asked if they would like to identify labels that do not match with \EBM’s classifications. They will also be asked to choose one of the two sampling methods. 

After the user clicks the Sample button, \EBM classifies all unlabeled records and suggests them as pseudo-labels. \XLabel then shows the pseudo-labels and the heatmaps (the explanations) in the main window. Regardless of the sampling method, records with low confidence scores will show up early during the labeling process. In the heatmaps, the red features are the main contributors to positive labels, while the blue features are those to negative labels. As we can see in Figure~\ref{fig:ss}, the main contributors of the positive label of Record \#6 are \texttt{DM\_key} and \texttt{DM\_ICD10}, both of which indicate that the patient have been diagnosed with diabetes mellitus (see the descriptions of these features in Table~\ref{features} below). 

Moreover, \XLabel can be used to detect mislabeled records, as shown in Figure~\ref{fig:ss}. We notice that Record \#6 was mislabeled as $0$, even though the features indicate that the label is $1$. \XLabel was able to identify the the label mismatch and suggest the correct label to the user. 

\section{Experiments} 

\subsection{Data description} 

Our dataset consists of the electronic health records of patient visits at two medical centers: one between February 1, 2022 to February 5, 2022, and the other on March 19, 2022. There might be multiple visits from the same patient within this period, in which case only one visit was randomly selected to ensure that the records are independent. Each record contains the patient's age, sex, height, weight, laboratory results, ICD-10 codes, prescribed drugs and the doctor's note. We asked a medical NCD expert to carefully read each record, and then append four binary labels, indicating the patient's status of four NCDs: diabetes mellitus (DM), hypertension (HTN), chronic kidney disease (CKD), and dyslipidemia (DLP).  The numbers of positive and negative records for each NCD are shown in Table~\ref{tabsum}.
\begin{table}
\caption{Number of records for each NCD\label{tabsum}}
\centering
\begin{tabularx}{0.95\textwidth}{lrrrr}
    \toprule
& Diabetes & Hypertension & Chronic Kidney Disease & Dyslipidemia \\ & (DM)	& (HTN)	& (CKD) & (DLP)\\
\midrule
  Positive & 72 & 139 & 52 & 77 \\
  Negative & 766 & 699 & 786 & 761 \\
  \bottomrule
  & & & Total & 838
\end{tabularx}
\end{table}

\subsection{Data Preprocessing}

It is inefficient to train \EBM on all features since most of the features are unrelated to a specific NCD type. Therefore, for each NCD type, we train \EBM only on a subset of features. The features suggested by the medical experts are listed in Table~\ref{features}. The complete list of keywords in the medical notes that are indicators of each NCD can be found in Table~\ref{tab5}.

\begin{table} 
\caption{List of input and label features for each NCD\label{features}}
\centering
\begin{tabularx}{\textwidth}{llX}
\toprule
NCD & Feature & Description   \\
\midrule
DM & \texttt{DM\_label} & 1 if diagnosed with diabetes, 0 otherwise \\
  & \texttt{DM\_key}     & 1 if the doctor's note has DM-related keywords, \\ & & 0 otherwise (see Table~\ref{tab5})      \\
   & \texttt{DM\_ICD10}     & 1 if at least one of the ICD-10 codes is DM-related, \\ & & 0 otherwise  \\
    & \texttt{DM\_drugs}     & 1 if the prescribed drugs are DM-related, 0 otherwise \\
    & \texttt{Glucose}   & Blood sugar level (mg/dL) \\ 
    & \texttt{HbA1c}     & Hemoglobin A1c (\%)                                  \\
    & \texttt{eGFR}     & Estimated glomerular filtration rate (mL/min/1.73m$^2$)                                 \\
    \midrule
    HTN & \texttt{HTN\_label} & 1 if diagnosed with hypertension, 0 otherwise \\
   & \texttt{HTN\_key}     & 1 if the doctor's note has HTN-related keywords,\\ & & 0 otherwise (see Table~\ref{tab5})      \\
   & \texttt{HTN\_ICD10}     & 1 if at least one of the ICD-10 codes is HTN-related, 0 otherwise \\
    & \texttt{HTN\_drugs}     & 1 if the prescribed drugs are HTN-related, 0 otherwise \\
    & \texttt{sbp1}   & Systolic blood pressure (mmHg) \\ 
    & \texttt{dbp1}     & Diastolic blood pressure (mmHg)                                 \\
\midrule
CKD & \texttt{CKD\_label} & 1 if diagnosed with chronic kidney disease, 0 otherwise \\
& \texttt{CKD\_key}     & 1 if the doctor's note has CKD-related keywords, \\ & & 0 otherwise (see Table~\ref{tab5})      \\
   & \texttt{CKD\_ICD10}     & 1 if at least one of the ICD-10 codes is CKD-related, 0 otherwise \\
    & \texttt{CKD\_drugs}     & 1 if the prescribed drugs are CKD-related, 0 otherwise \\
    & \texttt{DM\_pred}   & \EBM's classification of DM (0 or 1) \\ 
    & \texttt{HTN\_pred}     & \EBM's classification of HTN (0 or 1)                                \\
    & \texttt{eGFR}     & Estimated glomerular filtration rate (mL/min/1.73m$^2$)                                 \\
\midrule
DLP & \texttt{DLP\_label} & 1 if diagnosed with dyslipidemia, 0 otherwise   \\
& \texttt{DLP\_key}     & 1 if the doctor's note has DLP-related keywords, \\ & & 0 otherwise (see Table~\ref{tab5})      \\
   & \texttt{DLP\_ICD10}     & 1 if at least one of the ICD-10 codes is DLP-related,\\ & & 0 otherwise \\
    & \texttt{DLP\_drugs}     & 1 if the prescribed drugs are DLP-related, 0 otherwise \\
    & \texttt{Glucose}   & Blood sugar level (mg/dL) \\ 
    & \texttt{DM\_pred}   & \EBM's classification of DM (0 or 1) \\ 
    & \texttt{HTN\_pred}     & \EBM's classification of HTN (0 or 1)                                \\
    & \texttt{CKD\_pred}     & \EBM's classification of CKD (0 or 1)                                \\
    & \texttt{LDL-c}     & Low-density lipoprotein cholesterol (mg/dL)  \\
\bottomrule
\end{tabularx}
\end{table}
\begin{table}
\caption{Keywords associated with each NCD\label{tab5}}
\centering
\begin{tabularx}{0.68\textwidth}{llX}
\toprule
\textbf{NCD}	&\textbf{Feature name}  & \textbf{Keywords}\\
\midrule
DM	    & \texttt{DM\textunderscore key}   	& DM, diabetes, T1D, T2D\\
HTN	    & \texttt{HTN\textunderscore key}  	& HT, hypertension, bisoprolol\\
CKD	    & \texttt{CKD\textunderscore key}  	& CKD\\
DLP	    & \texttt{DLP\textunderscore key}  	& DLP, dyslipid, statin\\
\bottomrule
\end{tabularx}
\end{table}

Notice that the predictions for DM and HTN are input features to predict CKD and DLP, so the classifications have to be made in the following order: DM $\rightarrow$ HTN $\rightarrow$ CKD $\rightarrow$ DLP. 

\subsection{Details of the Experiments}

We shall perform three experiments to evaluate \XLabel and \EBM in terms of 1) number of user's labeling actions, 2) out-of-sample classification accuracy, and 3) label noise robustness.

\subsubsection*{Experiment 1: Evaluation of \XLabel.}
We evaluate \XLabel in terms of how much it helps with data labeling, measured by {\tt TotalFlips}, the total number of \XLabel's pseudo-labels that would have been corrected by the user over the whole dataset. For each NCD, we obtain an observed value of {\tt TotalFlips} by simulating the labeling process of \XLabel as follows:
\begin{enumerate}
  \item Start with the fully labeled dataset, with all labels hidden from \XLabel.
\item Randomly select 5\% of the records and reveal their labels to \XLabel.
\item Perform the following steps many times until all labels are revealed to \XLabel:
  \begin{itemize}
      \item  Train \EBM on the set of records whose labels have been revealed.
      \item Randomly select 20 of the remaining records and reveal their labels to \XLabel.
\item Use \EBM to predict the labels of those 20 records, count the number of incorrect predictions and add it to the number of labeling actions needed to be performed by the user.
  \end{itemize}
  \item After the labels of all records are revealed to \XLabel, let {\tt TotalFlips} be the numbers of labeling actions made by the user, which is the same as the number of incorrect predictions made by \EBM.
\end{enumerate}
This process gives us a single observed value {\tt TotalFlips} for each NCD. However, the successive performances of \EBM, and so the value of {\tt TotalFlips}, are greatly affected by the randomly initial 10 labels. To account for the randomness, we repeat the simulation 50 times with different initial labels and record the statistics of {\tt TotalFlips}.

We will compare \XLabel against a simple baseline model that classifies all NCDs of all records as negatives. If the user labeled the records with this baseline, the numbers of label corrections would be exactly the numbers of true positive labels as shown in Table~\ref{tabsum}. For \XLabel to outperform this baseline, the value of {\tt TotalFlips} for DM, HTN, CKD and DLP must be less than 72, 139, 52 and 77, respectively.

\subsubsection*{Experiment 2: \EBM's Classification Performance.} 

In the second experiment, we demonstrate that, in addition to being explainable, \EBM performs well compared to a baseline and several top-performing ML models for tabular data. Here, the baseline is a simple rule-based model (\ruleb) that classifies a medical record based on a well-known guideline for each disease. The full descriptions of \ruleb can be found in Appendix~\ref{sec:rulebased}. The ML models that are used to compare against \EBM are random forest (\RF), support vector machine (\SVM), implemented by~\cite{scikit-learn}, extreme gradient boosting machine (\XGB)~\citep{Chen2016} and light gradient boosting machine (\LGBM)~\citep{ke2017}. The hyperparameter settings of these models can be found in Appendix~\ref{sec:hyper}. 

To evaluate these models, we apply 5-fold cross-validation to assess their out-of-sample performances. The metrics that we use to evaluate the models are F1-score, Accuracy, Precision, and Recall. Among these, F1-score is our main performance metric, as it is robust to imbalanced data (e.g., a trivial model that classifies all records as 0 will receive a high accuracy but low F1-score). 

\subsubsection*{Experiment 3: \EBM's robustness to Label Noise.}

One purpose of \XLabel is to identify mislabeled records and correct them. To test \XLabel's capability in this regard, we perform an experiment to demonstrate that \EBM is robust to label noise, compared to the other models introduced in the previous experiment. 

This experiment starts with the dataset of electronic health records, which has been carefully labeled by medical experts. We flip the labels of a random sample consisting of $\{p\in5\%,10\%,…,50\%\}$ of the records. We then train \EBM on the dataset with the noisy labels and use it to classify the mislabeled records. The model’s robustness to label noise will be measured by how many of these classifications match with the true labels. In other words, we will measure the accuracy of the classifications on the mislabeled records.

We compare \EBM against \ruleb and several ML models. For each model and each percentage level $p$, we sample and train the models ten times and average the resulting accuracies. 

\section{Results and Discussion}

\subsection{Evaluation of \XLabel}
\begin{figure}[!t]
\centering\includegraphics[width=\textwidth]{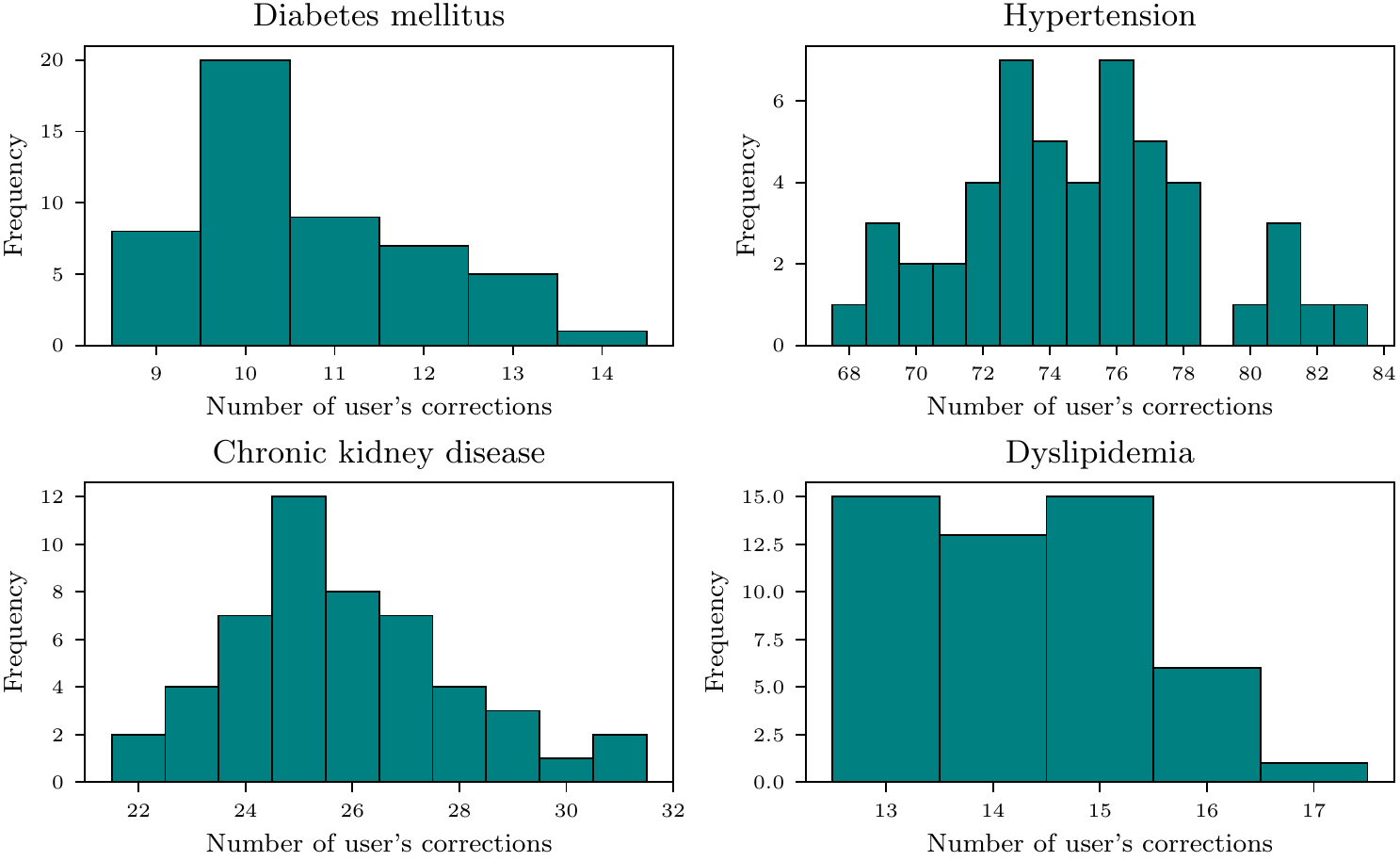}
\caption{The histograms of total number of label corrections ({\tt TotalFlips}) from 60 simulations of \XLabel's labeling process. \label{flips}}
\end{figure}   
\unskip

For each NCD, the histogram of 50 observed values of {\tt TotalFlips} is shown in Figure~\ref{flips}. Recall that our baseline is labeling every records as negative. In view of Table~\ref{tabsum}, this requires the user to change the 72, 139, 52 and 77 labels of DM, HTN, CKD and DLP, respectively. Based on Figure~\ref{flips}, it is evident that when using \XLabel, the user only needed to modify DM labels by one-fifth of the baseline, HTN labels by slightly more than half, CKD labels by approximately half, and DLP labels by one-fifth.

Nonetheless, there were some initial labels that resulted in relatively high numbers of label corrections. In particular, there are 6 initial labels of HTN that led to 74-77 label corrections. By inspecting these labels closely, we found that the poor results are caused by uninformative and homogeneous features. For example, sometimes a patient visited the medical center only to refill a prescription for hypertension medication. In this case, the only indicators of hypertension are ${\tt HTN\_key}=1$ and ${\tt HTN\_drugs} = 1$, while the other features are either $0$ or {\tt nan}; \EBM would learn nothing if the initial sample only consist of such records. Another example is when a HTN patient visited for treatment of non-HTN diseases. In this case, ${\tt HTN\_key}=0$ but ${\tt sbp1}$ and ${\tt dbp1}$ indicated that the patient had hypertension. If the initial sample contains many of such records, \EBM will incorrectly associate ${\tt HTN\_key}=1$ with ${\tt HTN\_label}=0$ at the start, and it will take many records to remove this association. From these observations, we conclude that in order to prevent a large number of label corrections, the first batch of labeled records must be sufficiently diverse in the input features.

\subsection{\EBM's classification performance} 

\begin{figure}[!t]
\centering\includegraphics[width=\textwidth]{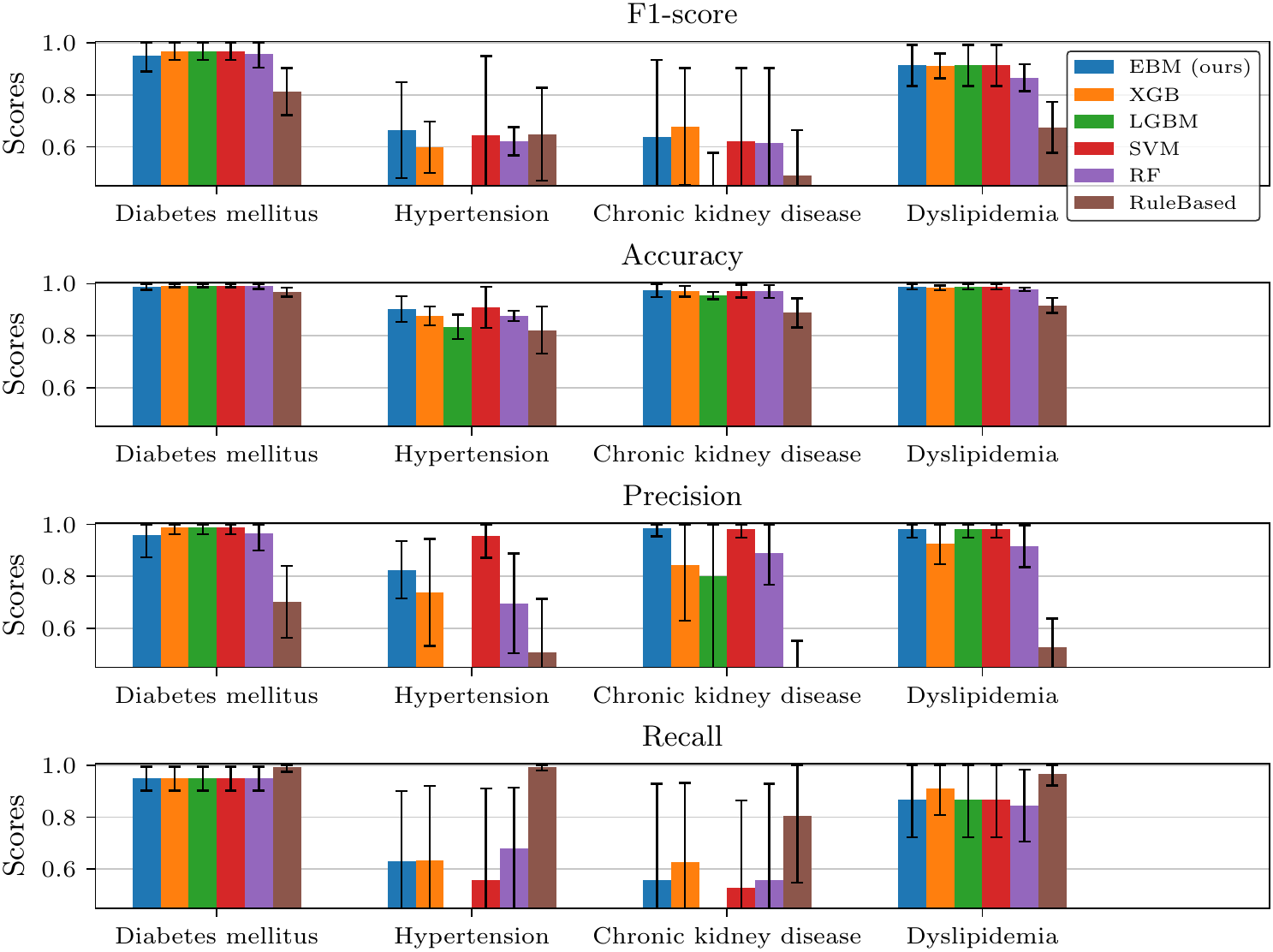}
\caption{The results of 5-fold cross-validations of \EBM, \SVM, \LGBM, \XGB, \RF and \ruleb. Here, four classification metrics across all NCDs are reported. The error bars represent the one standard deviations. \label{perf}}
\end{figure}   
\unskip

By performing the 5-fold cross-validation, we obtain five values for each metric, whose mean and standard deviation (SD) are reported as a bar chart with ±1SD error bars. The performances of \EBM and the other models for classifications of four NCDs are displayed by the metrics (rows) and the NCDs (columns) in Figure~\ref{perf}.

For DM classification, we see that \EBM performs as well as other ML models; sometimes it performs slightly worse, but all of its classification scores are still exceptionally high. For HTN, CKD and DLP classification, \EBM is always the best or the second best performers in terms of F1-score, accuracy and precision. We also notice that, while being outperformed by the other models in F1-score, accuracy, and precision, \ruleb has the highest recall rate in all NCDs, implying that the model is exceptional at identifying NCD patients, although with high false positive rates. \textbf{Since our goal is to reduce medical experts’ workload by making our label suggestions as accurate as possible, the ML models, which have higher F1-scores and accuracies, should be preferred over \ruleb}. 



To understand \EBM's misclassifications, we have inspected the records that it misclassified. Here are the records that we found: 
\begin{itemize}
     \item Records with typographical errors on important keywords in the doctor’s notes. Specifically, there is a medical note with ``DM'' mistyped as ``DN''. If the record contains no other indication of diabetes mellitus (such as high \texttt{glucose} or $\texttt{DM\_drugs}=1$), then \EBM will incorrectly classify such record as negative. 

    \item HTN-positive records with either $\texttt{HTN\_key}=0$, $\texttt{HTN\_ICD10}=0$ or $\texttt{HTN\_drugs}=0$ but the systolic or diastolic blood pressure is barely in the hypertension range; this is the case for patients who visited the medical center for non-HTN reasons. In our case, there are two patients with no indicator of HTN, while their blood pressures are 153/72 mmHg and 145/93 mmHg, respectively, which point to stage 2 hypertension.
\end{itemize}

The mistyping issue can be fixed with an edit distance-based spelling correction. For the second issue, \XLabel will give the record with a low confidence score. Such records will be brought to the user's attention early because of \XLabel's sampling methods. 

\subsection{\EBM's Robustness to Label Noise} 

\begin{figure}[!t]
\centering\includegraphics[width=\textwidth]{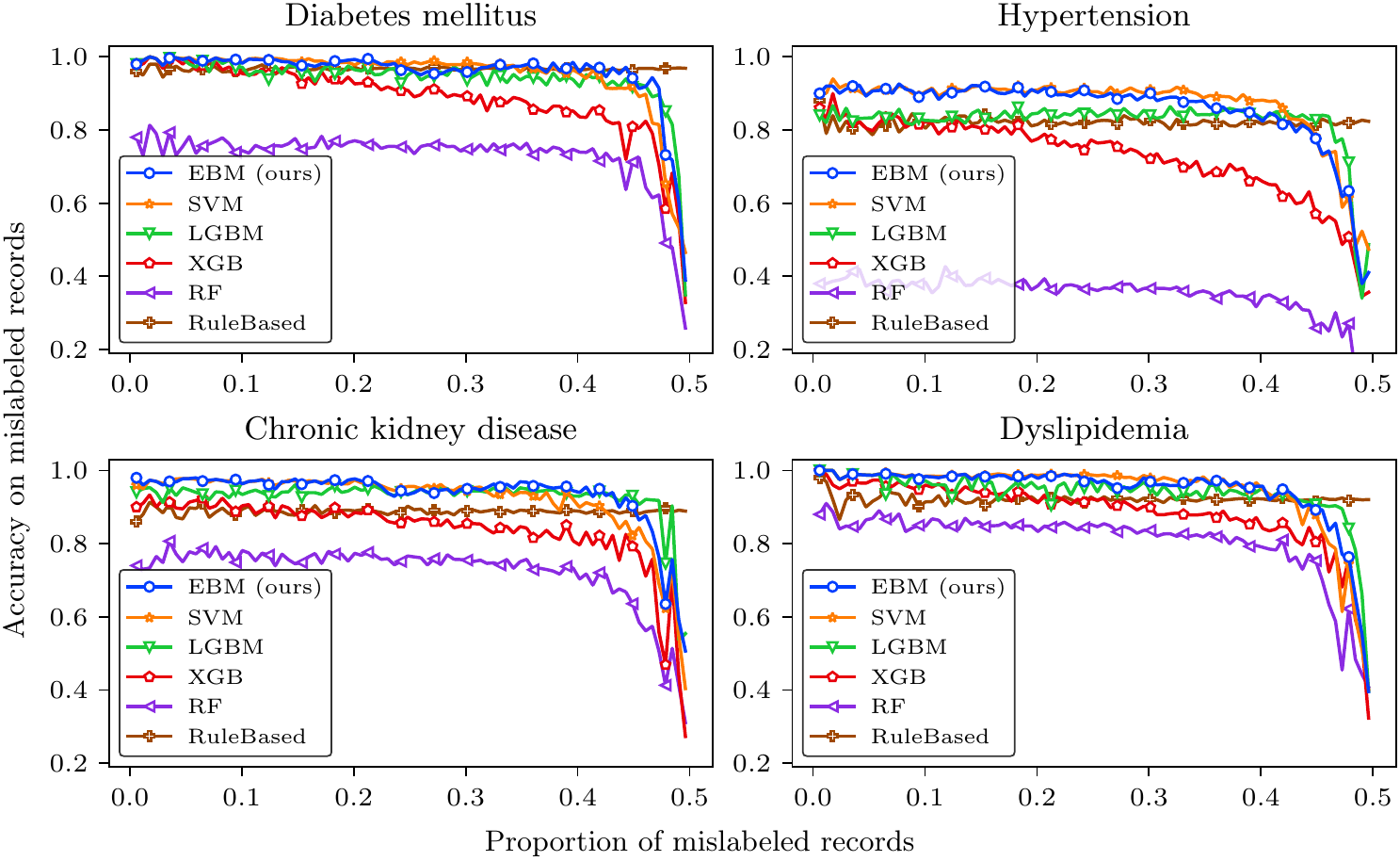}
\caption{The results of an experiment for label noise robustness of \EBM, \SVM, \LGBM, \XGB, \RF and \ruleb. In this experiment, we intentionally mislabeled a portion of the records, and then measured the accuracies of the models on the mislabeled data. Each of the four plots shows the accuracy ($y$-axis) against the proportion of mislabeled data ($x$-axis) for each NCD. \label{LN}}
\end{figure}   
\unskip

The plots of the models’ average accuracies against the proportion of mislabeled records for each NCD are shown in Figure~\ref{LN}. From the figure, we see that \EBM, \SVM, \LGBM, and \ruleb are the most robust to label noises. We note that the accuracy of \ruleb does not go down with the label noise, because \ruleb is a set of rules based solely on the features, and not the label. However, when only 5\%–20\% of the records are mislabeled (or 5\%–45\% in the cases of HTN, CKD and DLP), it is less accurate than \EBM, \SVM, and \LGBM. 

Even when 40\%–45\% of the records are mislabeled, \EBM, \SVM, and \LGBM still retain their high accuracies; in the cases of diabetes mellitus, hypertension, and dyslipidemia, \EBM is more accurate than the other ML models. Although the accuracies of these models drop quickly after the 45\% mark, we do not expect this many records to be mislabeled in a real-life scenario. 

\section{Conclusions}

We developed Explainable Labeling Assistant (\XLabel), a new visual-interactive tool for electronic health record labeling. The main feature of \XLabel is its ability to suggest the labels of new records, as well as the explanations in the form of heatmaps of the contributions of the input features. At a high level, \XLabel trains and employs Explainable Boosting Machine (\EBM) to classify the records' labels. Our first experiment shows that \XLabel helps reduce the number of users' actions per one labeling session over the whole dataset. The second experiment shows that \EBM is a good choice of explainable classification model as it outperforms a rule-based model used by medical experts, and performs on par with popular gradient boosting models. And the third experiment shows that \EBM is very robust to label noise; even when $40\%$ of the data are mislabeled, \EBM can recall almost all of the true labels. Even though \XLabel was employed specifically to label NCD data, we hope that \XLabel will be of use in other labeling tasks as well.

\section*{Acknowledgements}

We thank Sriphat Medical Center, Chiang Mai, Thailand for providing valuable data. We also thank the physician team for labeling and validating data accuracy.

\section*{Disclosure statement}

The authors report there are no competing interests to declare.

\section*{Funding}

This work was supported by Fundamental Fund 2022, Chiang Mai University under grant number FF65/059.

\section*{Data availability statement}
Due to the nature of the research and ethical restrictions, supporting data is not available.







\bibliographystyle{tfcse}
\bibliography{NCDLabel}

\begin{thebibliography}{31}
\providecommand{\natexlab}[1]{#1}
\providecommand{\url}[1]{\normalfont{#1}}
\providecommand{\urlprefix}{Available from: }

\bibitem[Ambika et~al.(2020{\natexlab{a}})]{Ambika2020a}
Ambika~M, Raghuraman~G, SaiRamesh~L. 2020{\natexlab{a}}. Enhanced decision
  support system to predict and prevent hypertension using computational
  intelligence techniques. Soft Computing. 24(17):13293--13304.
  \urlprefix\url{https://doi.org/10.1007/s00500-020-04743-9}.

\bibitem[Ambika et~al.(2020{\natexlab{b}})]{Ambika2020b}
Ambika~M, Raghuraman~G, SaiRamesh~L, Ayyasamy~A. 2020{\natexlab{b}}.
  Intelligence -- based decision support system for diagnosing the incidence of
  hypertensive type. Journal of Intelligent {\&} Fuzzy Systems. 38:1811--1825.
  2;  \urlprefix\url{https://doi.org/10.3233/JIFS-190143}.

\bibitem[Apley and Zhu(2020)]{Daniel2020}
Apley~DW, Zhu~J. 2020. {Visualizing the effects of predictor variables in black
  box supervised learning models}. Journal of the Royal Statistical Society
  Series B. 82(4):1059--1086.
  \urlprefix\url{https://ideas.repec.org/a/bla/jorssb/v82y2020i4p1059-1086.html}.

\bibitem[Ashktorab et~al.(2021)]{Ashktorab2021}
Ashktorab~Z, Desmond~M, Andres~J, Muller~M, Joshi~NN, Brachman~M, Sharma~A,
  Brimijoin~K, Pan~Q, Wolf~CT, et~al. 2021. Ai-assisted human labeling:
  Batching for efficiency without overreliance. Proc ACM Hum-Comput Interact.
  5(CSCW1).  \urlprefix\url{https://doi.org/10.1145/3449163}.

\bibitem[Bondi et~al.(2022)]{Bondi2022}
Bondi~E, Koster~R, Sheahan~H, Chadwick~M, Bachrach~Y, Cemgil~T, Paquet~U,
  Dvijotham~K. 2022. Role of human-ai interaction in selective prediction.
  Proceedings of the AAAI Conference on Artificial Intelligence.
  36(5):5286--5294.
  \urlprefix\url{https://ojs.aaai.org/index.php/AAAI/article/view/20465}.

\bibitem[Chen et~al.(2019)]{Chen2019}
Chen~H, Lundberg~SM, Lee~S. 2019. Explaining models by propagating shapley
  values of local components. CoRR. abs/1911.11888.
  \urlprefix\url{http://arxiv.org/abs/1911.11888}.

\bibitem[Chen and Guestrin(2016)]{Chen2016}
Chen~T, Guestrin~C. 2016. {XGBoost}: A scalable tree boosting system. In:
  Proceedings of the 22nd ACM SIGKDD International Conference on Knowledge
  Discovery and Data Mining; New York, NY, USA. ACM. p. 785--794. KDD '16;
  \urlprefix\url{http://doi.acm.org/10.1145/2939672.2939785}.

\bibitem[Cheng et~al.(2020)]{cheng2020}
Cheng~D, Ting~C, Ho~C, Ho~C. 2020. Performance evaluation of explainable
  machine learning on non-communicable diseases. Solid State Technol.
  63:2780--2793.

\bibitem[Davagdorj et~al.(2021)]{Davagdorj2021}
Davagdorj~K, Bae~JW, Pham~VH, Theera-Umpon~N, Ryu~KH. 2021. Explainable
  artificial intelligence based framework for non-communicable diseases
  prediction. IEEE Access. 9:123672--123688.

\bibitem[Desmond et~al.(2021)]{Desmond2021}
Desmond~M, Muller~M, Ashktorab~Z, Dugan~C, Duesterwald~E, Brimijoin~K,
  Finegan-Dollak~C, Brachman~M, Sharma~A, Joshi~NN, et~al. 2021. Increasing the
  speed and accuracy of data labeling through an ai assisted interface. In:
  26th International Conference on Intelligent User Interfaces; New York, NY,
  USA. Association for Computing Machinery. p. 392–401. IUI '21;
  \urlprefix\url{https://doi.org/10.1145/3397481.3450698}.

\bibitem[Finkelstein and Jeong(2017)]{Finkelstein2017}
Finkelstein~J, Jeong~Ic. 2017. Machine learning approaches to personalize early
  prediction of asthma exacerbations. Annals of the New York Academy of
  Sciences. 1387(1):153--165.
  \urlprefix\url{https://nyaspubs.onlinelibrary.wiley.com/doi/abs/10.1111/nyas.13218}.

\bibitem[Friedman(2001)]{Friedman2001}
Friedman~JH. 2001. Greedy function approximation: A gradient boosting machine.
  The Annals of Statistics. 29(5):1189--1232.  [accessed 2022-08-22].
  \urlprefix\url{http://www.jstor.org/stable/2699986}.

\bibitem[Friedman(2002)]{Friedman2002}
Friedman~JH. 2002. Stochastic gradient boosting. Computational Statistics \&
  Data Analysis. 38(4):367--378. Nonlinear Methods and Data Mining;
  \urlprefix\url{https://www.sciencedirect.com/science/article/pii/S0167947301000652}.

\bibitem[Islam et~al.(2020)]{islam2020b}
Islam~MT, Raihan~M, Farzana~F, Aktar~N, Ghosh~P, Kabiraj~S. 2020. Typical and
  non-typical diabetes disease prediction using random forest algorithm. In:
  2020 11th International Conference on Computing, Communication and Networking
  Technologies (ICCCNT). p. 1--6.

\bibitem[Ke et~al.(2017)]{ke2017}
Ke~G, Meng~Q, Finley~T, Wang~T, Chen~W, Ma~W, Ye~Q, Liu~TY. 2017. Lightgbm: A
  highly efficient gradient boosting decision tree. Advances in neural
  information processing systems. 30:3146--3154.

\bibitem[Lipton(2018)]{Lipton2018}
Lipton~ZC. 2018. The mythos of model interpretability: In machine learning, the
  concept of interpretability is both important and slippery. Queue.
  16(3):31–57.  \urlprefix\url{https://doi.org/10.1145/3236386.3241340}.

\bibitem[Lou et~al.(2013)]{Lou2013}
Lou~Y, Caruana~R, Gehrke~J, Hooker~G. 2013. Accurate intelligible models with
  pairwise interactions. In: Proceedings of the 19th ACM SIGKDD International
  Conference on Knowledge Discovery and Data Mining; New York, NY, USA.
  Association for Computing Machinery. p. 623–631. KDD '13;
  \urlprefix\url{https://doi.org/10.1145/2487575.2487579}.

\bibitem[Lundberg and Lee(2017)]{Lundberg2017}
Lundberg~SM, Lee~SI. 2017. A unified approach to interpreting model
  predictions. In: Guyon~I, Luxburg~UV, Bengio~S, Wallach~H, Fergus~R,
  Vishwanathan~S, Garnett~R, editors. Advances in Neural Information Processing
  Systems; vol.~30. Curran Associates, Inc.
  \urlprefix\url{https://proceedings.neurips.cc/paper/2017/file/8a20a8621978632d76c43dfd28b67767-Paper.pdf}.

\bibitem[Mali and Singh(2022)]{mali2022}
Mali~B, Singh~PK. 2022. Towards simulating a global robust model for early
  asthma detection. In: Phillipson~F, Eichler~G, Erfurth~C, Fahrnberger~G,
  editors. Innovations for Community Services; Cham. Springer International
  Publishing. p. 257--266.

\bibitem[Nadj et~al.(2020)]{Nadj2020}
Nadj~M, Knaeble~M, Li~MX, Maedche~A. 2020. Power to the oracle? design
  principles for interactive labeling systems in machine learning. {KI} -
  K\"{u}nstliche Intelligenz. 34(2):131--142.
  \urlprefix\url{https://doi.org/10.1007/s13218-020-00634-1}.

\bibitem[Nori et~al.(2019)]{nori2019}
Nori~H, Jenkins~S, Koch~P, Caruana~R. 2019. Interpretml: A unified framework
  for machine learning interpretability. arXiv preprint arXiv:190909223.

\bibitem[Pedregosa et~al.(2011)]{scikit-learn}
Pedregosa~F, Varoquaux~G, Gramfort~A, Michel~V, Thirion~B, Grisel~O, Blondel~M,
  Prettenhofer~P, Weiss~R, Dubourg~V, et~al. 2011. Scikit-learn: Machine
  learning in {P}ython. Journal of Machine Learning Research. 12:2825--2830.

\bibitem[Pei et~al.(2019)]{Pei2019}
Pei~D, Gong~Y, Kang~H, Zhang~C, Guo~Q. 2019. Accurate and rapid screening model
  for potential diabetes mellitus. BMC Medical Informatics and Decision Making.
  19(1):41.  \urlprefix\url{https://doi.org/10.1186/s12911-019-0790-3}.

\bibitem[Rajora et~al.(2021)]{Rajora2021}
Rajora~M, Rathod~M, Naik~NS. 2021. Stroke prediction using machine learning in
  a distributed environment. In: Goswami~D, Hoang~TA, editors. Distributed
  Computing and Internet Technology; Cham. Springer International Publishing.
  p. 238--252.

\bibitem[Rashed-Al-Mahfuz et~al.(2021)]{Rashed2021}
Rashed-Al-Mahfuz~M, Haque~A, Azad~A, Alyami~SA, Quinn~JMW, Moni~MA. 2021.
  Clinically applicable machine learning approaches to identify attributes of
  chronic kidney disease (ckd) for use in low-cost diagnostic screening. IEEE
  Journal of Translational Engineering in Health and Medicine. 9:1--11.

\bibitem[Ribeiro et~al.(2018)]{Ribeiro2018}
Ribeiro~MT, Singh~S, Guestrin~C. 2018. Anchors: High-precision model-agnostic
  explanations. Proceedings of the AAAI Conference on Artificial Intelligence.
  32(1).
  \urlprefix\url{https://ojs.aaai.org/index.php/AAAI/article/view/11491}.

\bibitem[Rosado and Hernandez(2019)]{rosado2019}
Rosado~JT, Hernandez~AA. 2019. Developing a predictive model of stroke using
  support vector machine. In: 2019 IEEE 13th International Conference on
  Telecommunication Systems, Services, and Applications (TSSA). p. 35--40.

\bibitem[Rudin(2019)]{Rudin2019}
Rudin~C. 2019. Stop explaining black box machine learning models for high
  stakes decisions and use interpretable models instead. Nature Machine
  Intelligence. 1(5):206--215.
  \urlprefix\url{https://doi.org/10.1038/s42256-019-0048-x}.

\bibitem[Shafi et~al.(2022)]{Shafi2022}
Shafi~J, Basu~S, Kavila~SD. 2022. Role of explainable artificial intelligence
  ({XAI}) in prediction of non-communicable diseases ({NCDs}). In: Advances in
  medical technologies and clinical practice. {IGI} Global; p. 113--130.
  \urlprefix\url{https://doi.org/10.4018/978-1-6684-3791-9.ch005}.

\bibitem[Viana et~al.(2021)]{Viana2021}
Viana~L, Oliveira~E, Conte~T. 2021. An interface design catalog for interactive
  labeling systems. In: Proceedings of the 23rd International Conference on
  Enterprise Information Systems. {SCITEPRESS} - Science and Technology
  Publications.  \urlprefix\url{https://doi.org/10.5220/0010458204830494}.

\bibitem[Yakimovich et~al.(2021)]{YAKIMOVICH2021100383}
Yakimovich~A, Beaugnon~A, Huang~Y, Ozkirimli~E. 2021. Labels in a haystack:
  Approaches beyond supervised learning in biomedical applications. Patterns.
  2(12):100383.
  \urlprefix\url{https://www.sciencedirect.com/science/article/pii/S2666389921002506}.

\end{thebibliography}

\appendix

\section{Rule-based Classification Models}\label{sec:rulebased}
\begin{figure}
\centering\includegraphics[width=\textwidth]{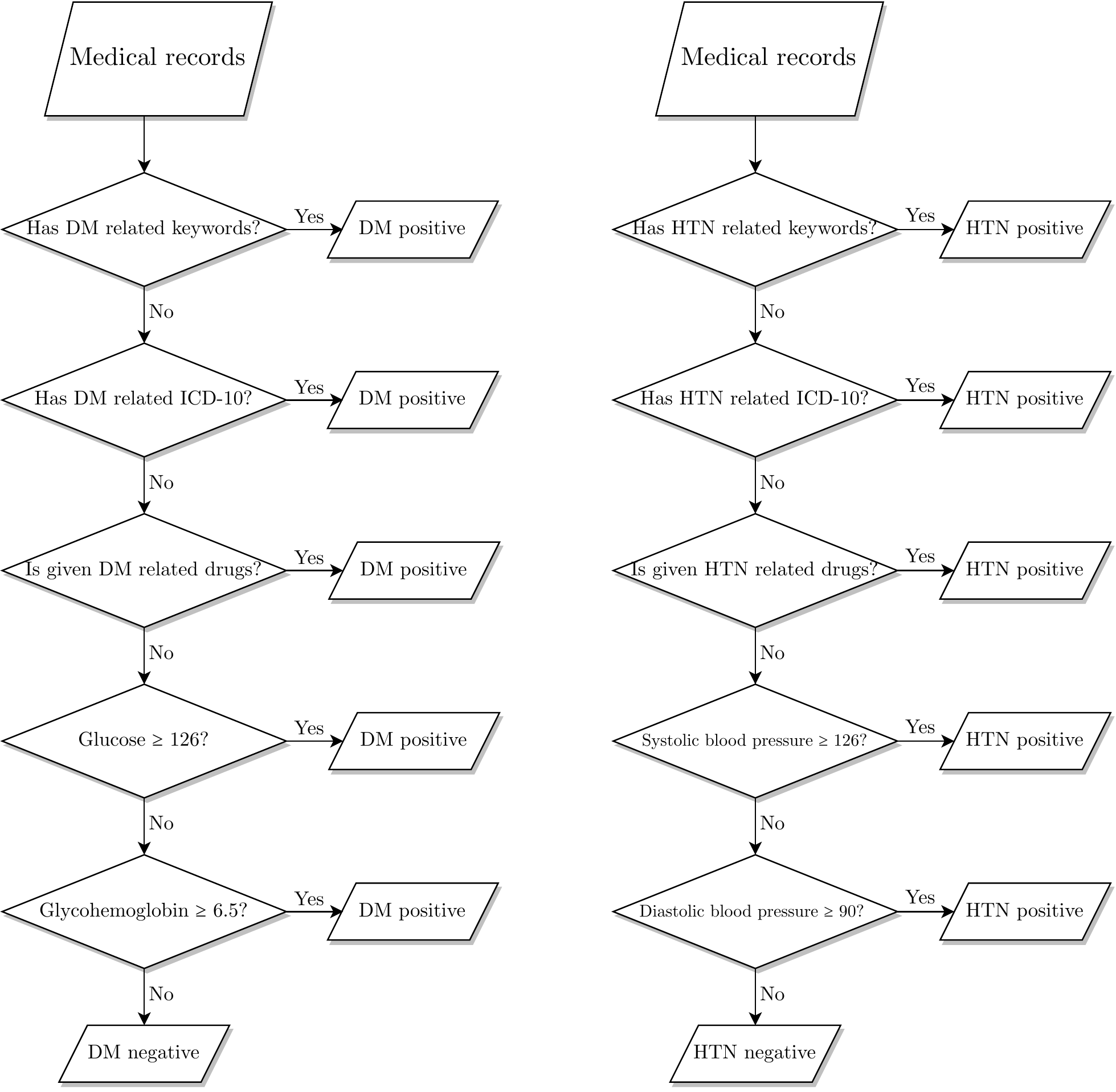}
\caption{Left: a flow chart of \ruleb for DM classification. Right: a flow chart of \ruleb for HTN classification.  \label{RB1}}
\end{figure}

\begin{figure}
\centering\includegraphics[width=\textwidth]{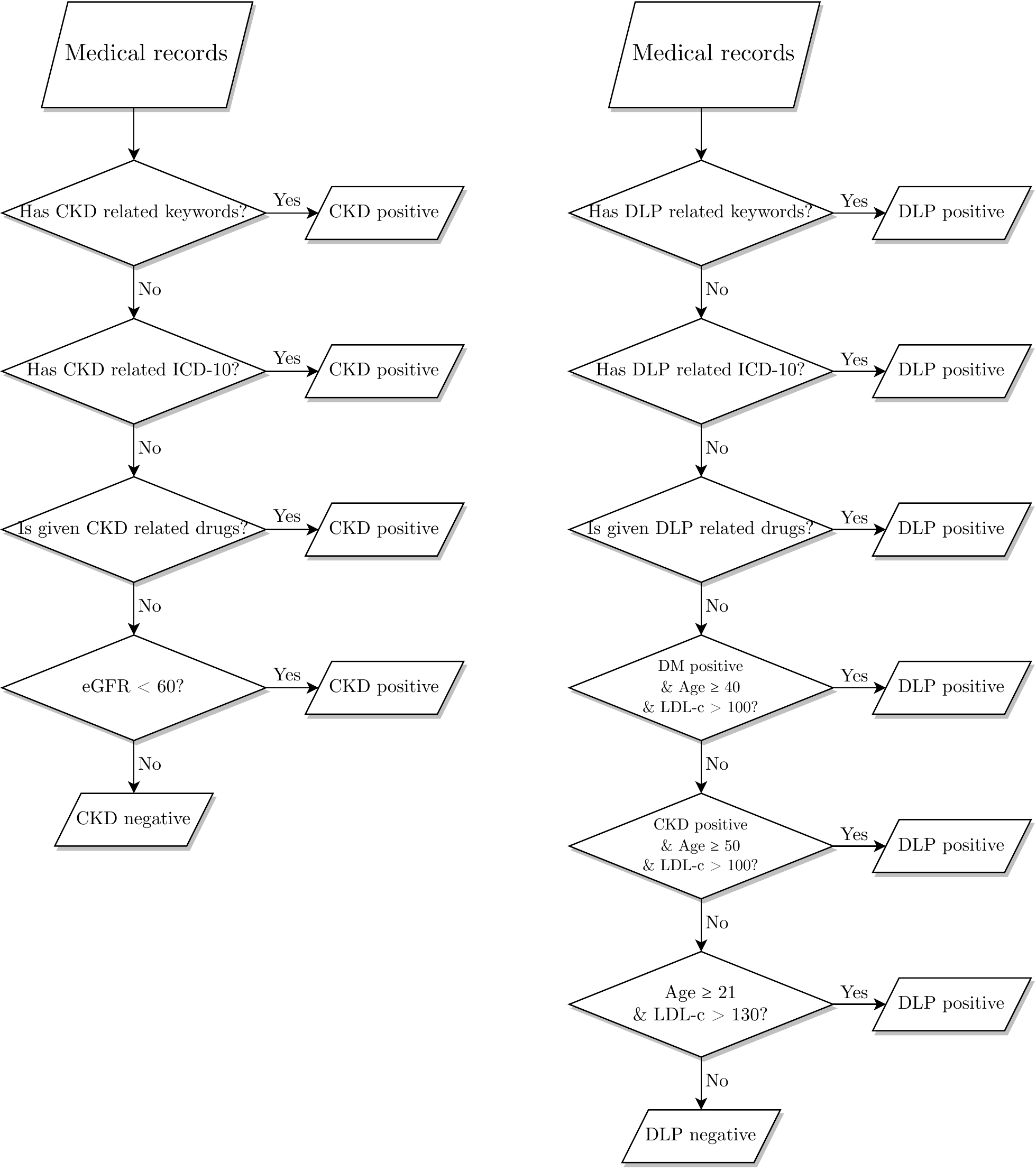}
\caption{Left: a flow chart of \ruleb for CKD classification. Right: a flow chart of \ruleb for DLP classification. \label{RB2}}
\end{figure}

The details of \ruleb for classification of diabetes mellitus (DM), hypertension (HTN), chronic kidney disease (CKD) and dyslipidemia (DLP) are shown in Figure~\ref{RB1} and Figure~\ref{RB2}.

\newpage

\section{Models' hyperparameters}\label{sec:hyper}
Most of the hyperparameters are set at the corresponding package's default values. All hyperparameters with non-default values are shown in Table~\ref{hyper}.
\begin{table}[!h] 
\caption{None-default hyperparameters of the machine learning models \label{hyper}}
\begin{tabularx}{\textwidth}{lllX}
\toprule
Model & Hyperparameter                                                 & NCD & Value                              \\
\midrule
\RF    & Number of trees    & DM                                                              & 200 \\
& & HTN, DLP & 100 \\
& & CKD & 50 \\
      & Minimum number of records per leaf     & All & 1                                  \\
\midrule
\XGB   & Number of trees                                                               & All    & 2                                  \\  
& Minimum sum of record weight in a leaf &  DM & 1 \\
 & & HTN, CKD, DLP & 0          \\
\midrule
\LGBM  & Number of trees   & DM, DLP                                                                & 6 \\
& & HTN, CKD & 5          \\  
      & Minimum number of records per leaf      & All & 1                                  \\ 
      & Maximum number of leaves  & All & 2                                  \\
\midrule
\SVM   & Soft margin constant (C) & DM, DLP & 0.1  \\
& & HTN, CKD & 0.05     \\  
      & Kernel             & All                                                                & Linear                             \\
      & Optimization & All                                                                      & Primal                             \\
\midrule
\EBM   & Number of pairwise interactions                                                         & All             & 0                                  \\
      & Maximum number of bins in feature binning                                                             & All             & 3                                 \\
\bottomrule
\end{tabularx}
\end{table}

\end{document}